# FUSION OF REAL TIME THERMAL IMAGE AND 1D/2D/3D DEPTH LASER READINGS FOR REMOTE THERMAL SENSING IN INDUSTRIAL PLANTS BY MEANS OF UAVs and/or ROBOTS[1]

Corneliu Arsene

Department of Electrical and Electronic Engineering, University of Manchester,
Manchester, UK
arsenecorneliu@tutanota.de

## Abstract

This paper presents fast procedures for thermal infrared remote sensing in dark, GPS-denied environments, such as those found in industrial plants such as in High-Voltage Direct Current (HVDC) converter stations. These procedures are based on the combination of the depth estimation obtained from either a 1-Dimensional LIDAR laser or a 2-Dimensional Hokuyo laser or a 3D MultiSense SLB laser sensor and the visible and thermal cameras from a FLIR Duo R dual-sensor thermal camera. The combination of these sensors/cameras is suitable to be mounted on Unmanned Aerial Vehicles (UAVs) and/or robots in order to provide reliable information about the potential malfunctions, which can be found within the hazardous environment. For example, the capabilities of the developed software and hardware system corresponding to the combination of the 1-D LIDAR sensor and the Flir Duo R dual-sensor thermal camera are assessed from the point of the accuracy of results and the required computational times: the obtained computational times are under 10 ms, with a maximum localization error of 8 mm and an average standard deviation for the measured temperatures of 1.11°C, which results are obtained for a number of test cases.

The paper is structured as follows: the description of the system used for identification and localization of hotspots in industrial plants is presented in section II. In section III, the method for faults identification and localization in plants by using a 1-Dimensional LIDAR laser sensor and thermal images is described together with results. In section IV the real time thermal image processing is presented. Fusion of the 2-Dimensional depth laser Hokuyo and the thermal images is described in section V. In section VI the combination of the 3D MultiSense SLB laser and thermal images is described. In section VII a discussion and several conclusions are drawn.

---

[1] Any suggestions and comments are welcome at arsenecorneliu@tutanota.de

# 1. INTRODUCTION

This paper addresses the fault identification challenge found in various industrial plants such as HVDC offshore electrical stations. The faults [1] may appear because for example of the overheating of the electrical connectors in converter valve equipment. A significant challenge is the ability to gather sensor data whilst the plant is 'live'. Installing thermal sensors throughout the infrastructure may be logistically infeasible due to the number of sensors required and the management of the additional data streams. An alternative solution which is being explored is the use of robots and/or Unmanned Aerial Vehicles (UAVs), which if especially built (on purpose) then might be able to operate in the presence of high electromagnetic fields [2-4]. In [5] it was described the architecture of a system, which once mounted on UAVs, it would enable autonomous or tele-operation of the drone within a dark and GPS-denied environment. It consisted of 3D estimation of the drone by using sensor fusion [6] and an Extended Kalman Filter [7] for the estimation of dynamic state variable represented by the 6 Degrees of Freedom (DoF) of the drone. Quick Response (QR) codes [8] would also be used to indicate that the drone is near a specific point of the valve racks in a HVDC electrical station. This system would be now enhanced with two more functionalities for the identification of the overheated valves (i.e. hotspots) and the precise localization of these hotspots (Fig.1). In fact by looking to Fig. 1, it is possible to assume that the coordinates of the hotspot position would be the UAV coordinates plus the displacement $z_h$ but this would be true as long as $x_h$ and $y_h$ are zero, which is not always the case.

The paper is structured as follows: the description of the system used for identification and localization of hotspots is presented in section II. In section III, the method for faults identification and localization in plants by using a 1-Dimensional LIDAR laser sensor and thermal images is described together with results. In section IV the real time thermal image processing is presented. Fusion of the 2-Dimensional depth laser Hokuyo and the thermal images is described in section V. In section VI the combination of the 3D MultiSense SLB laser and thermal images is described. In section VII a discussion and several conclusions are drawn.

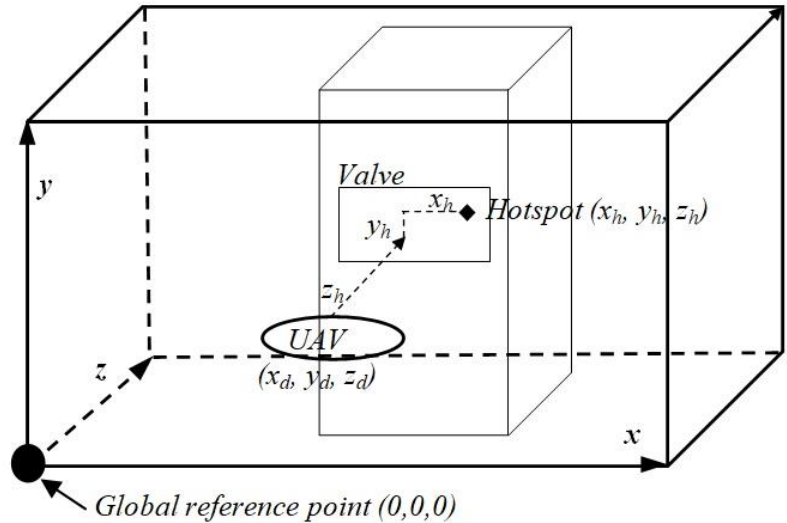

**Fig. 1.** Industrial plant with position of UAV ($x_d$, $y_d$, $z_d$), converter valve and increased temperatures for a valve (i.e. hotspot localization) ($x_h$, $y_h$, $z_h$).

## 2. SYSTEM OVERVIEW

Visual odometry and mapping for autonomous flight using RGB-Depth camera [9] has captured the attention of robotics researchers for long time. A system made of multiple drones and sensors for building inspection was developed in [10]. In [11] is shown a method developed for creating 3D thermal models using a thermal camera and a Microsoft Kinect depth camera [12]. In [13] a thermal and a visual camera were used in order to perform the 3D thermal reconstruction and based on the acquisition of two sets of two images of the same object but from two different points of view [14]. In [15-20] there are also discussed combination of thermal images and visible images for 3D thermal model reconstruction although the entire computational times even for small scenes could be high (i.e. more than 1 minute). A combination [21] of a 3D laser with an RGB camera and a thermal camera was employed to devise 3D thermal point cloud models of interior buildings but which was not installed on an autonomous system of navigation or any other robot. More generally [22] the importance at a larger scale of thermal imaging for detection of hotspots for streets and neighborhoods was also envisaged by using airborne remotely sense data but again difficulties for example with costs of obtaining and processing such images were envisaged.

Very recently [23-27] fusion of 2D and 3D laser and thermal sensors have been developed and presented successfully for various applications but by using non-aerial robots. However, these approaches may require an increased number of sensors/cameras which would be difficult to be mounted on a UAV because of lack of space and the additional weight. In this context, the aim here is to advance the technological know-how and to develop more simplified and fast hardware and software systems for detection and localization in real-time of hotspots in industrial plants by using especially UAVs and/or robots.

The developed algorithms are implemented first for testing purposes in Matlab and Arduino C/C++ and then they are transferred by the mean of OpenCV library running on to an Ubuntu operating system installed on an Intel® NUC8 with a $8^{th}$ generation Intel Core™ processor with 32GB DDR4 RAM. The other component used it is a LIDAR-Lite v3 laser ranging module (Fig.2a) with a 40 meter laser-based sensor. The communication can be done via I2C and/or PWM communication protocols. The LIDAR laser will provide the depth information to the identified hotspots.

A dual thermal sensor will be used, which is a FLIR Duo R dual-sensor thermal camera (Fig.2b), which has two cameras: a visible camera and a thermal infrared camera. For simplicity we may refer from here onwards also as to the Flir Duo R but the actual Flir Duo R dual-sensor thermal camera has as already described two cameras, a visible camera and a thermal infrared camera.

The thermal infrared camera senses and images long wave infrared radiation with an uncooled V0x (Vanadium Oxide) microbolometer. It has thermal measurement accuracy of 5°C and a sensor resolution of 160x120 pixels. The benefit of using this Flir Duo R is that the visible camera offers the possibility of making visual inspections while the thermal camera offers the thermal information: it has also a reduced size 41x59x29.6 mm and it weighs only 84 grams, which makes it suitable to be mounted on a UAV. The visible camera resolution is 1920x1080 pixels.

The digital video output consists of a Micro-HDMI port displaying HDMI videos at 1080p and both visible and thermal images. The thermal and visible images can be read in Matlab software under Windows operating system by using the FLIR drivers, which are called Atlas® drivers. Furthermore, with a video capture card it is possible to stream in real time thermal/visible videos for example to an Ubuntu operating system. The laser device and the thermal sensor are connected and communicating to the Intel®NUC8 (Fig.2c).

The combination of all the previous described electronics items are mounted on a DJI MATRICE M210 V2 drone (Fig.2d), which has a maximum payload of 1.34 kg corresponding to a 24 minutes maximum flying time.

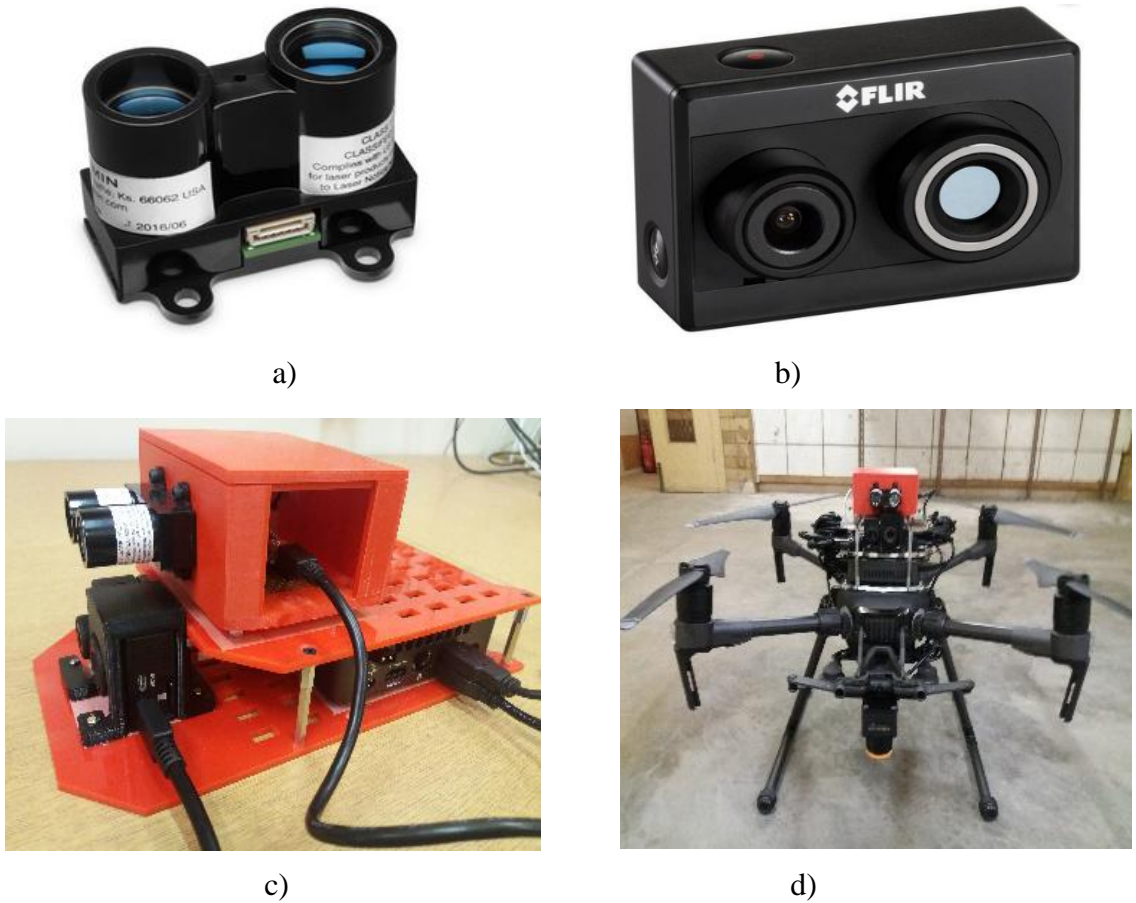

**Fig. 2.** Software and hardware system: a) LIDAR laser; b) FLIR Duo R; c) Connecting LIDAR and FLIR Duo R to the Intel® NUC8; d) DJI MATRICE M210 V2 drone with sensors installed on it.

The use of visible and thermal cameras requires an initial calibration. Calibration (i.e. calculation of intrinsic and extrinsic camera parameters) of visible camera requires a calibration pattern such as a chessboard pattern. However, the calibration of thermal infrared cameras is more difficult as it requires that the calibration pattern to somehow emit heat. There have been used various methods: in [28], a printed calibration pattern (i.e. chessboard) was used and placed on a glazed finish ceramic tile backing to maintain the pattern flat and to keep the heat when using a heat lamp. Another approach [29] has been to print a chessboard pattern onto a manufactured Printed Circuit Board (PCB), which would be heated with a fan heater. A special fabricated mask was presented in [30] where the pattern was held in front of a monitor to be detected by the thermal camera. In order to calibrate multi-spectral imaging system and under different environment condition, a robust chessboard pattern [33] was made of a combination of sandblast (i.e. minimizes specular reflections) and aluminum (i.e. extends contrast persistence). It was also attempted [31] joint calibration of visible and far-infrared cameras with reported pixel errors of proposed system smaller than other existing systems. Joint calibration of thermal and visible cameras was also made in [32] for the purposes of 3D mapping of object surface temperatures. In this work, it will be used a paper chessboard pattern for the visible camera and a PCB pattern [28] for the thermal camera.

## 3. FAULT IDENTIFICATION AND LOCALIZATION

As described previously, this paper presents a fast procedure for fault identification and localization which can be used in industrial plants by using UAV or robots. The combination of sensors installed on the drone tries to detect the overheated regions on valves (i.e. hotspots) (Fig.3) by using the pixel information from the visible and thermal images obtained with the FLIR Duo R dual-sensor. In Fig.3, it is shown an example of a visible image (i.e. lamp, Fig.3a) (.jpg image, resolution 1080x1920x3 uint8) obtained with the visible camera and the corresponding thermal image (Fig.3b) obtained with the thermal camera of FLIR Duo R dual-sensor that is .tiff image, array size 120x160. The second picture (Fig.3b) is completely white because of how is produced by the thermal camera, that is pixel values are higher than the pixel value of 255, which corresponds to the white color. In Fig.3c the .tiff image file (i.e. displayed with imagesc function from Matlab) consists of an array/matrix of size 120x160, and by selecting a point in the .tiff image, it is shown a hotspot temperature of 44.15°C. It was determined from Fig.3b by multiplying a value from Fig.3b such as 31730 (i.e. value produced by Flir Duo R dual-sensor thermal camera) with 0.01 and subtracting 273.15: this is how the raw values (e.g. 31730) provided by the Flir Duo R dual-sensor thermal camera have to be processed in order to obtain the temperature in degrees Celsius (i.e. by multiplying 0.01 and subtracting 273.15). The radiometric jpeg picture (Fig.3d) is obtained by FLIR Duo R by combining the visible with the thermal information from both types of pictures (i.e. .jpg image, resolution 480x640x3 uint8). The radiometric picture has associated a matrix of temperatures, of size 240x320 in Celsius degrees, which matrix of temperatures is encoded into the radiometric picture. Normally, the temperature obtained in Fig.3c at a given point should be similar with the temperature in Fig.3d at the same location. The four pictures from Fig.3 are showing similar information and since the resolutions are known, it is possible to calculate the pixel coordinates of the same point through the four pictures. Furthermore, the localization of the hotspots is based on the depth information acquired from LIDAR laser. Supposing that the UAV position ($x_d$, $y_d$, $z_d$) is known with regard to a global reference point (0,0,0), by using the navigation system described in [5] and the depth information $d$ ($z_h = d$) available from the 1D laser LIDAR, then it is possible to calculate the other coordinates $y_h$ and $x_h$ of the hotspot with regard to the position of the camera as follows (i.e. the pinhole camera model). It is assumed in this work for simplicity that the camera coordinate system is the same as the world and the hotspots coordinate system, and therefore the extrinsic camera matrix becomes the identity matrix:

$$x_h = d * (x_{pixel} - c_x) / f_x$$
$$y_h = d * (y_{pixel} - c_y) / f_y \qquad (1)$$
$$z_h = d$$

where $x_{pixel}$, $y_{pixel}$ are the pixel coordinates in the visible picture, $f_x$, $f_y$ are the focal lengths (pixels), $c_x$, $c_y$ are the coordinates of the optical center of the image (pixels).

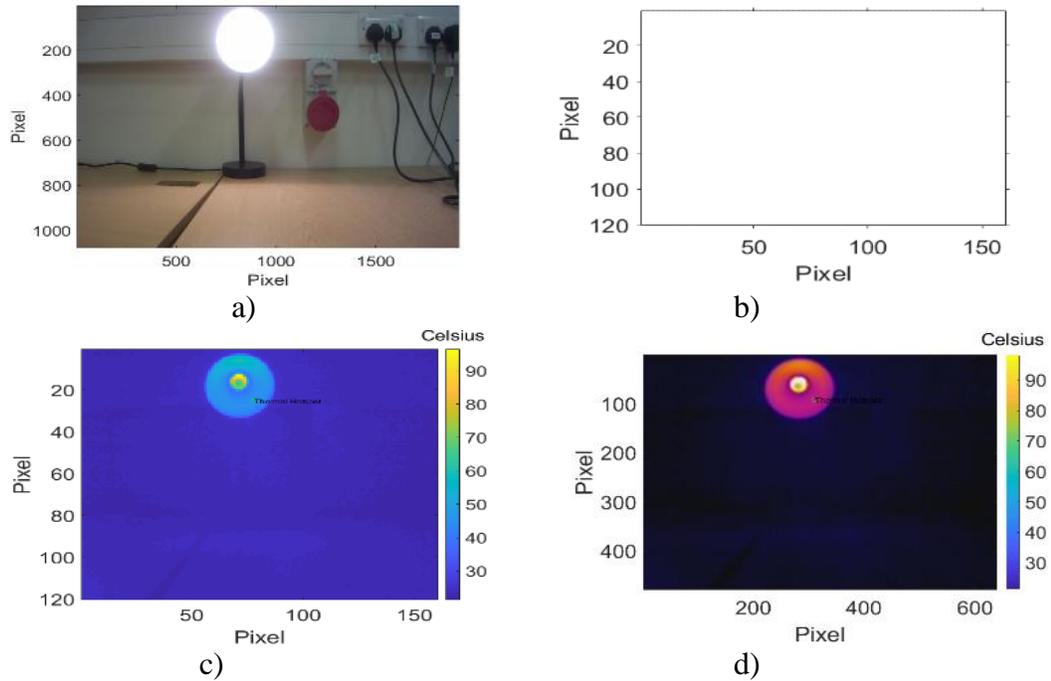

**Fig.3.** FLIR Duo R dual-sensor: a) visible picture (.jpeg); b) original thermal picture as taken by the thermal camera (.tiff) (i.e. white image); c) extracted thermal picture from the (b)-original thermal camera picture; d) radiometric jpeg picture combining the visual and the thermal information (R.jpeg).

The focal lengths and the optical center coordinates form the intrinsic parameters $K_1$ of the visible camera and it is shown below. $K_1$ is obtained through the calibration of the camera and as already explained, it is easier to calibrate and use the intrinsic parameters of the visible camera than to calibrate and use the intrinsic parameters of a thermal camera:

$$K_1 = \begin{bmatrix} f_x & 0 & 0 \\ 0 & f_y & 0 \\ c_x & c_y & 1 \end{bmatrix} \qquad (2)$$

The hotspot position with regard to the FLIR Duo R dual-sensor is determined with eq. (1). By using the assumption that the rotation matrix of a UAV with regard to the global reference point is the identity matrix, then the global position of hotspot with regard to the global reference point (Fig. 1) could be simply written as:

$$x_f = x_d + x_h$$
$$y_f = y_d + y_h \qquad (3)$$
$$z_f = z_d + z_h$$

where $x_f$, $y_f$ and $z_f$ are the final positions of the hotspot calculated with regard to the global reference point (Fig.1).

Eqs. (1) to (3) are valid as long as the laser beam is perpendicular to the hotspot plane (Fig.1). However, if mounted on a drone, then the angles of rotations of the drone will have to be taken into consideration and the equations will have to be updated. In Fig.4 is shown the central position (mm) of a blob hotspot with regard to the thermal camera corresponding to Fig.3c and characterized by temperatures higher than 44.45°C and coordinates (-64.19, 109.06, 710.00). The computational times to calculate the coordinates are obviously very small and are given by the acquisition of the depth measurements by the LIDAR laser (i.e. 3.62 ms) and the calculations from eq. (1) (i.e. 3 ms), which in total is under 10 ms.

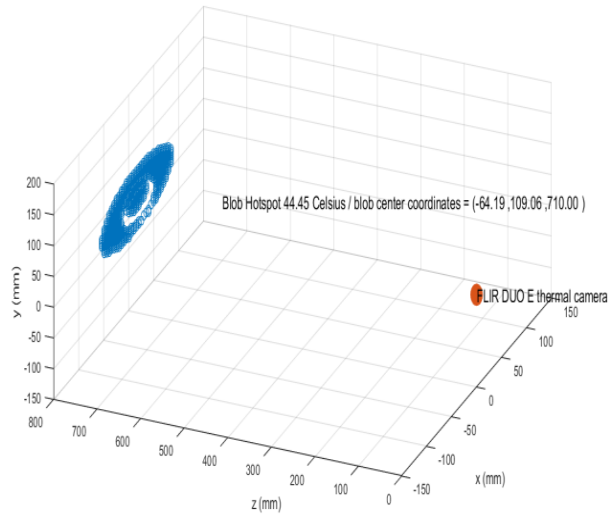

**Fig.4.** Blob hotspot localization with regard to the position of the Flir Duo R dual-sensor.

As a test case scenario, the method is further applied on a PCB chessboard (Fig.5a), which was used also for calibrating the thermal camera that in the end is mounted on the drone (Fig.2d). There are two cases: the heated PCB is located at shorter and then at longer distances (depth) from the thermal sensor and 10 thermal measurements are taken for the centre and the lower left corner (Table 1). The distances are chosen to test the situations which can appear for example inside a HVDC electrical substation where the maximum distance between two valve towers is usually 5 m (i.e. 5000 mm). For the short distance (i.e. 858.8 mm), for the centre square colored in white (i.e. measured from left - $7^{th}$ horizontally and measured from up - $4^{th}$ vertically) the mean temperature was 52.74°C with standard deviation ±1.30°C (Table 1). In the lower left corner for the white square the mean temperature was 40.70°C with standard deviation ±0.36°C. The difference is due to the heat fan blowing behind the chessboard towards the centre area. The laser took in less than a minute 1400 consecutive distance measurements with a mean value of 858.8 mm and standard deviation of ±8.1 mm and therefore within the manufacturer error bounds ±25 mm. Then, the PCB chessboard was placed at a long distance of 4075.6 mm (Fig.5c). The temperatures for the squares situated at the centre and at the margin are more similar and uninform across the entire PCB chessboard: mean temperature of 44.81°C and standard deviation of ±0.28°C for the centre square and mean temperature of 43.67°C and standard deviation of ± 2.51°C for the margin square. This is explained by the long distance of over 4 meters from the thermal sensor and that the thermal radiation is weak at such distance. The laser took within a minute 1400 distance measurements with mean value of 4075.6 mm and standard deviation (std) of ± 8.0 mm, which is within the error bounds reported by the manufacturer of ±25 mm. Table 2 shows how errors (i.e. ±8.0 mm) affect the x, y, z coordinates.

The described system would be able to successfully identify hotspots inside an industrial plant while being mounted on UAVs or robot.

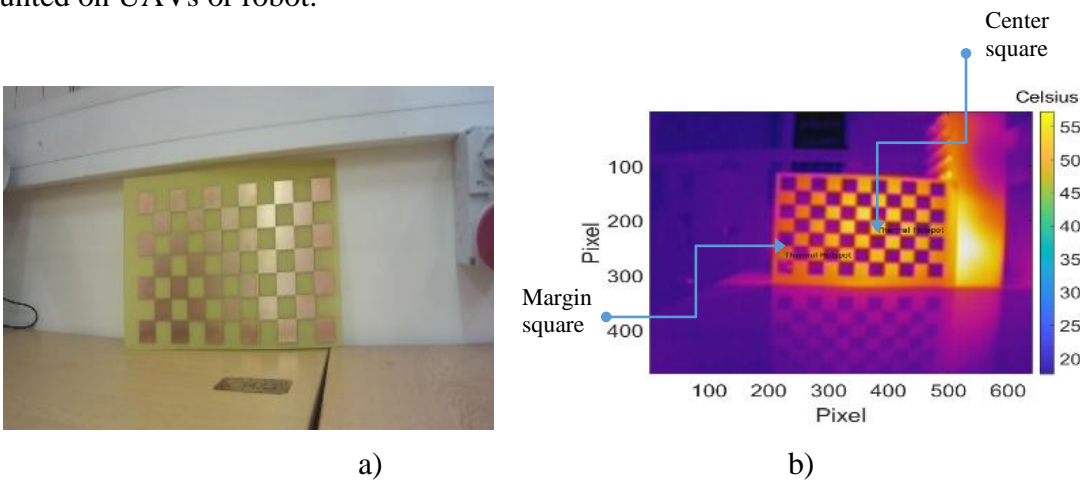

a)  b)

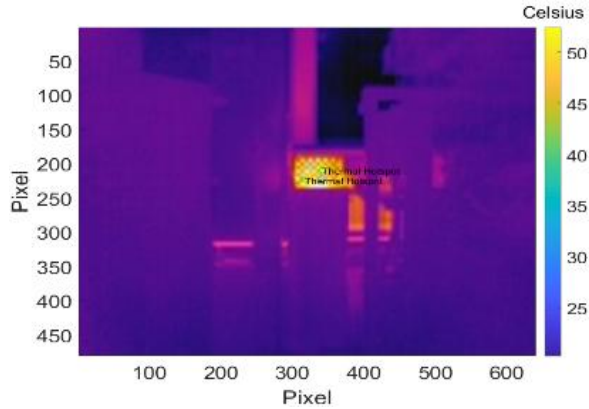

c)

**Fig.5.** PCB chessboard with squares made of cooper: a) PCB; b) heated chessboard PCB (i.e. radiometric jpeg image) located at short distance of 1D LIDAR laser; c) heated chessboard PCB located at longer distance from 1D LIDAR laser.

| Distance | Centre square | Margin square |
|---|---|---|
| Short   858.8 mm | 52.74 ± 1.30°C | 40.70 ± 0.36°C |
| Long 4075.6  mm | 44.81± 0.28°C | 43.67 ± 2.51°C |

**Table 1.** Temperatures at short and long distances from the thermal sensor in two different locations: centre and left low margin square of PCB chessboard pattern.

| Position | **X** (mm) | **Y** (mm) | **Z** (mm) |
|---|---|---|---|
| Mean | 15.17 | -174.22 | 4075.6 |
| Mean+Std | 15.1670 | -174.2182 | 4083.6 |

**Table 2.** X, Y, Z coordinates affected by small errors.

## 4.  REAL TIME THERMAL IMAGE PROCESSING

For a robot or a UAV (e.g drone), which is equipped with a thermal camera, it is paramount to be able to access the thermal sensor (e.g. FLIR Duo R dual-sensor) in real time in order to access the latest taken thermal image.

This can be done programmatically in C/C++ from a NUC computer running Linux/Ubuntu operating system and located on top of the UAV or on a robot. It can be also done at specific moments of time or by running a loop where at each number of seconds/minutes (e.g. 10 seconds, 15 seconds, 1 minute, 2 minutes etc) the operation can take place. Assuming that the thermal camera is identified by the NUC's Linux/Ubuntu operating system as being with the ID '3237-3231' then the latest thermal image is always spotted at the end of running the below software procedure:

```
if ((dir == opendir ("/media/nuc/3237-3231/" != NULL ) {
    /* print all files and directories within directory */
    while (( ent = readdir (dir)) != NULL)  {
        sprintf (buff, "%s" , ent->d_name);
    }
    closedir (dir);
}
```

**Fig.6.** Typical software procedure for reading directories/files stored on Flir Duo R dual-sensor.

In Fig.6 dir (DIR), ent (dirent), buff (stat) are C/C++ variables of type DIR, dirent and stat, which are used by the software procedure from above for reading directories/files stored on FLIR Duo R dual-sensor.

For the visual image from Fig.7a (resolution 1080x1920x3), the corresponding thermal image taken in real time and saved on the FLIR Duo R dual-sensor thermal camera is shown in Fig. 7b. In real time and by using software procedures similar to the one from Fig.6 and C/C++ functions from OpenCV, the raw values from Fig7b are multiplied with 0.01 and are subtracted by 273.15 and the reason for doing this mathematical operation is because of the pixel picture format provided by the Flir Duo R dual-sensor thermal camera. This will produce the real temperature data in Celsius degrees saved as the gray picture from Fig.7c (resolution 120x160). In Fig.7d is shown Fig.7c after applying a color map (e.g. COLORMAP_RAINBOW) with the C/C++ OpenCV function applyColorMap. Although at the first glance figures Fig. 7b to Fig.7d seem to be uniform, by inspecting a bit more (i.e. zooming might be needed) it can be noticed slightly different variations in the colors of the images, which correspond to warmer or cooler areas in the thermal images.

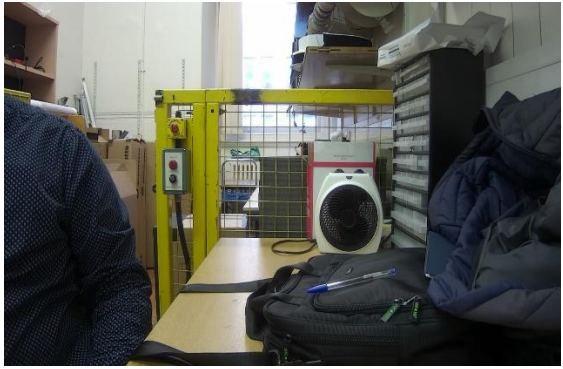

a)

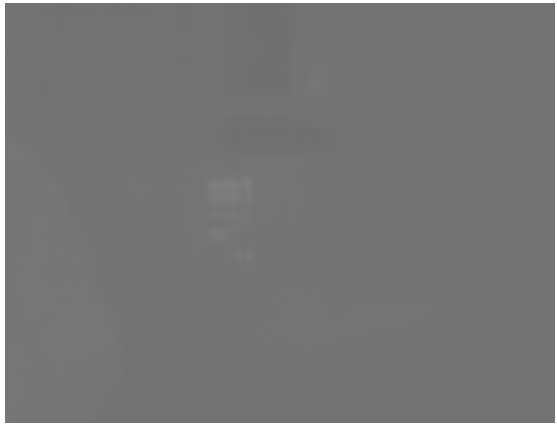

b)

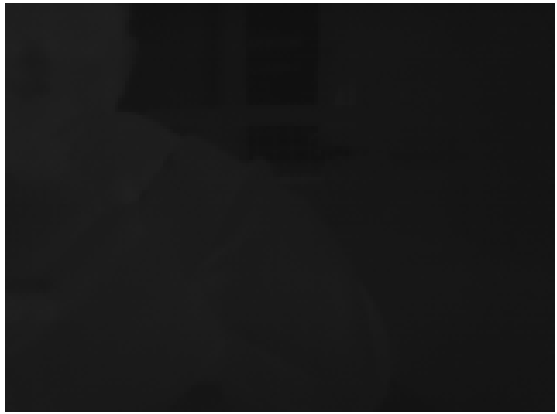

c)

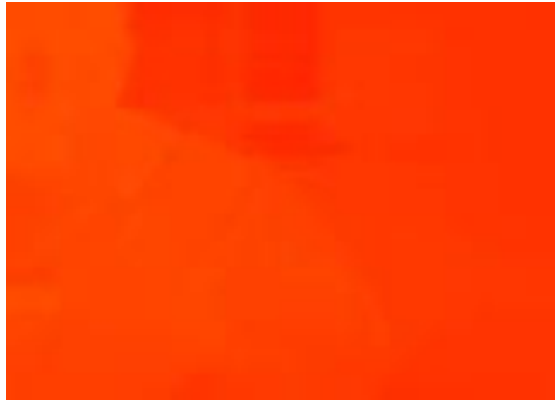

d)

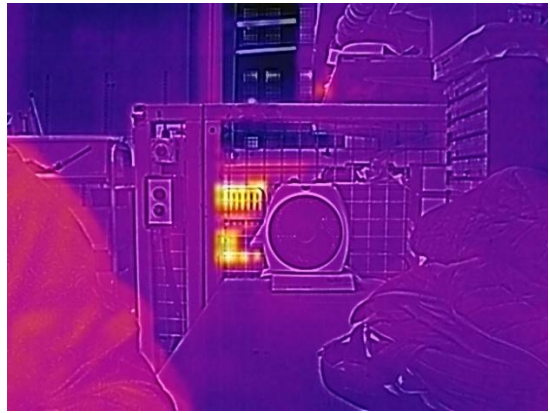

e)

**Fig.7.** Real time temperature data extraction from thermal images: a) visual image; b) thermal image in .tiff format; c) extracted Celsius degrees temperatures and saved in a .jpeg file; d) applying a colormap in OpenCV on c) image; e) MSX enhanced radiometric jpeg thermal image as taken by the Flir Duo R dual-sensor thermal camera.

Finally, Fig.7e (resolution 480x640x3) shows the MSX (Multi-Spectral Dynamic Imaging) enhanced radiometric thermal images. The FLIR MSX technology adds visible light details to the thermal images in real time for higher clarity of the radiometric thermal images. A Robot Operating System (ROS) [34] node has been also written in C/C++ programming language (Ubuntu/Linux operating system) which publishes thermal video camera messages that can be visualized with RViz [35], which is a 3D visualization tool for ROS.

Once the temperature in Celsius degrees has been obtained (Fig.7c) in real time, the temperature at a point in space is given by the following equation and by using the imread function from the OpenCV image processing (C/C++ programming language), which is usually used for reading images stored on the local hard drives:

$$T_{Celsius} = \text{Thermal\_image}_{Celsius}(x_{pixel}, y_{pixel}) \quad (4)$$

where $x_{pixel}$, $y_{pixel}$ are the pixel coordinates in the Celsius thermal image from Fig.7c.

However eq. (4) requires the pixel coordinates ($x_{pixel}$, $y_{pixel}$) which are obtained from an inverse of a system of equations similar to the one shown in equation (1):

$$x_{pixel} = c_x + f_x * (x_h / d)$$
$$y_{pixel} = c_y + f_y * (y_h / d) \quad (5)$$

where $x_h$, $y_h$, $z_h$ are the coordinates in space of a point where the temperature is needed, $d$ is the depth information (i.e. distance from the thermal camera to a point in space).

The coordinates of a point in space ($x_h$, $y_h$, $z_h$) can be obtained by using either a 1D, 2D or 3D laser sensor. In the previous section (i.e. section III) the coordinates were obtained from a 1D LIDAR laser sensor while in the following sections the coordinates will be obtained from a 2D Hokuyo laser and a 3D MultiSense SLB laser.

## 5. FUSION OF 2D DEPTH LASER DATA AND THERMAL IMAGE

For measuring the distances to the hotspots it can be used also a 2D dimensional Hokuyo laser (UST-10LX) [36] (Fig. 8), which has 1081 measurement steps, 270° the detection angle, 0.25° the angular resolution and the scanning direction from Fig.9a. This section will present several theoretical and practical results by using also the RViz software application which is a 3D visualization tool for ROS and under Ubuntu operating system.

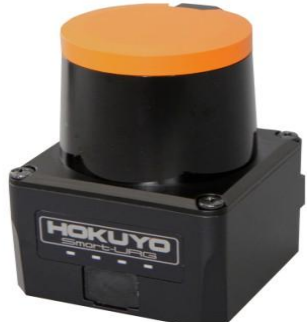

**Fig. 8**. Hokuyo 2D laser.

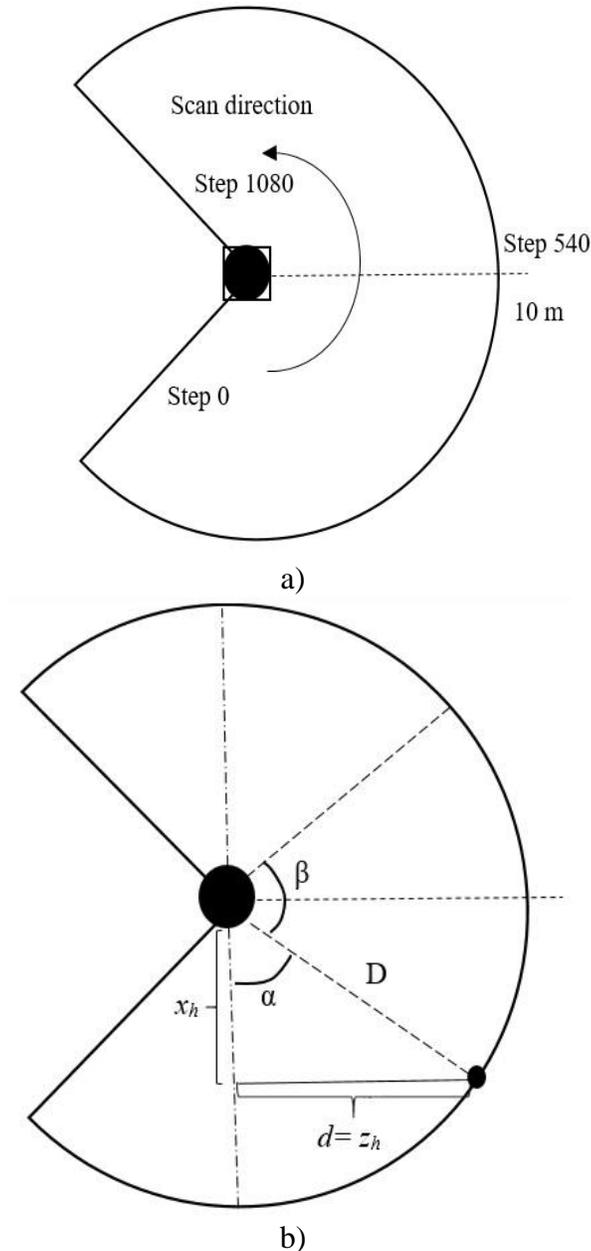

**Fig.9.** Determining the 3D position of a point in space with a 2D Hokuyo laser: a) the way the 2D laser works; b) the determination of the 3D position of a point in space.

In Fig.9b there are shown the α angle measured by the 2D laser sensor, β is the field of view of the thermal camera measured to be 60 degrees, D is the known distance to the point in space measured by the 2D laser and $d$ ($z_h$) is the depth information denoted similarly as in eq.(5). This is true if it is assumed that the incidence of the laser beam and the thermal camera are perpendicular to the plane of interest. By knowing α and D from the 2D laser sensor then it is possible to determine the other variables of interest:

$$x_h = D \cos(\alpha)$$
$$d = D \sin(\alpha) \qquad (6)$$

Variable $y_h$ represents rather a constant given by the displacement (i.e. on vertical axis) of the 2D laser with respect to the thermal camera of the Flir Duo R dual-sensor as shown in Fig.10. FOV stands for Field of View of the thermal camera and we are interested in the 2D laser readings which are within the field of view of the thermal camera.

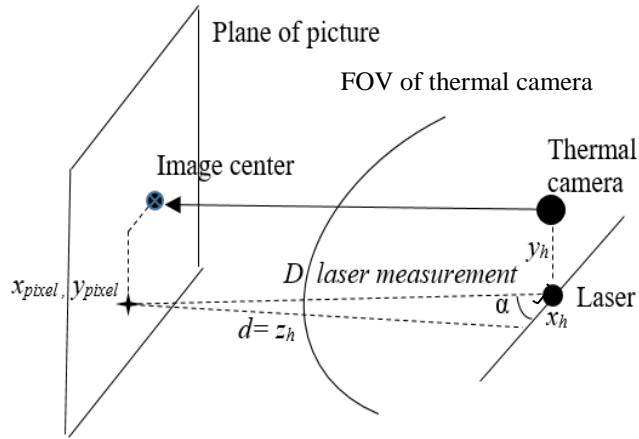

**Fig.10.** Combination of the 2D laser and thermal camera measurements.

Once $x_h$, $y_h$, $d$ are known then $x_{pixel}$, $y_{pixel}$ can be determined with eq. (5). An urg_node ROS is used, which is suitable for the 2D dimensional Hokuyo laser (UST-10LX). It publishes the laser points and it is launched with the command rosrun urg_node urg_node _ip_address:=192.168.0.10.

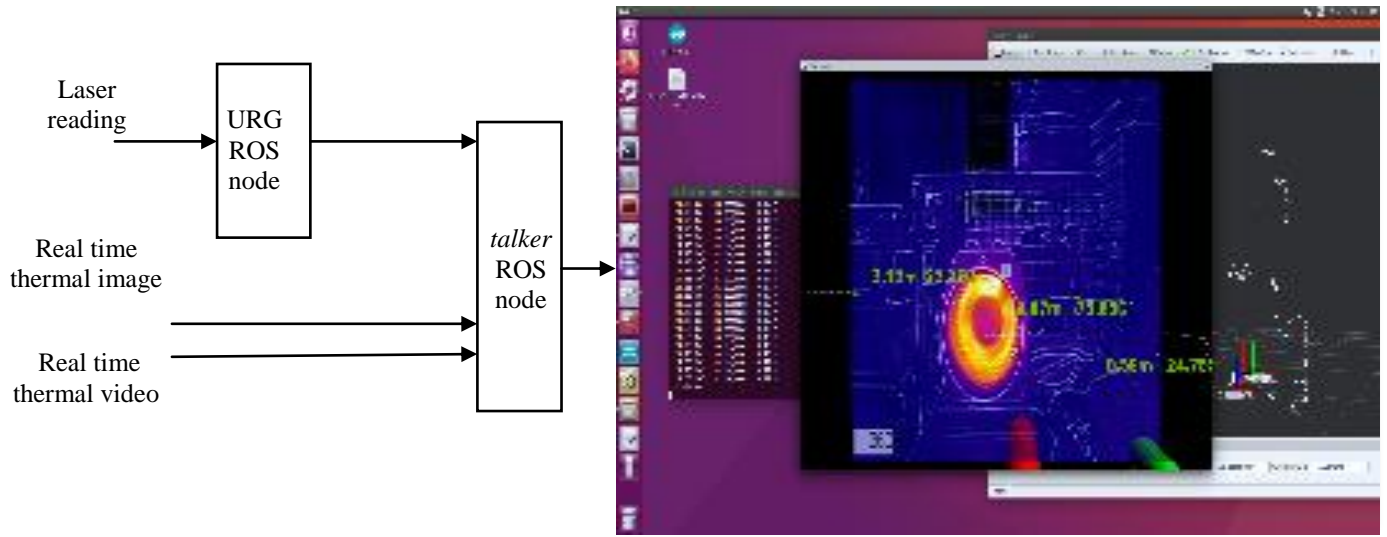

Publishing information on RViz

**Fig.11.** ROS nodes for processing of 2D laser readings, real time thermal video and real time thermal image for hotspots identification and localization.

A different node (e.g. called "talker") subscribes to the above urg_node and processes the laser readings by using the eqs. (4-6). In the same time, in real time, temperatures at some given locations in the 3D space can be obtained continuously from the 14 bit .tiff files as explained above in section 4. The flow of information and processing is depicted in Fig.11 whereas it can be seen the RViz module opened behind the camera screen capture.

The real-time information is published in RViz as shown in Fig. 11. There are chosen 3 points, the left and the right outermost points and a central location situated under the small green box. For the central green box the temperature is provided automatically from fabric by the thermal camera in the lower left corner of the real time thermal image/video.

For each of these 3 points, every time when a new temperature is sought, a new thermal picture has to be taken. One way to do this is by using an Arduino board and connecting it to the thermal camera. The Pulse Width Modulated (PWM) protocol can be used from the Arduino board to the thermal camera to take one or multiple thermal images or even to start recording multi-page tiff files containing the raw sensor data (i.e. temperatures). At the other end, the Arduino board would be connected to the NUC computer situated on the UAV and the Arduino board would receive commands of when to take the thermal pictures. This could happen periodically such as each 10 seconds or at some specific moments of time when the human operator decides. The *talker* ROS node (Fig.11) is of interest as it publishes two pieces of information: the real time thermal video together with the three laser points and their associated information that is the distances and the temperatures in Celsius degrees. It is possible to publish a larger number of points function of the decision of the human operator. For information purposes, the URG ROS node is also allowed to publish independently the laser readings as some interrupted very small white lines over the real time thermal video (i.e. Fig.12).

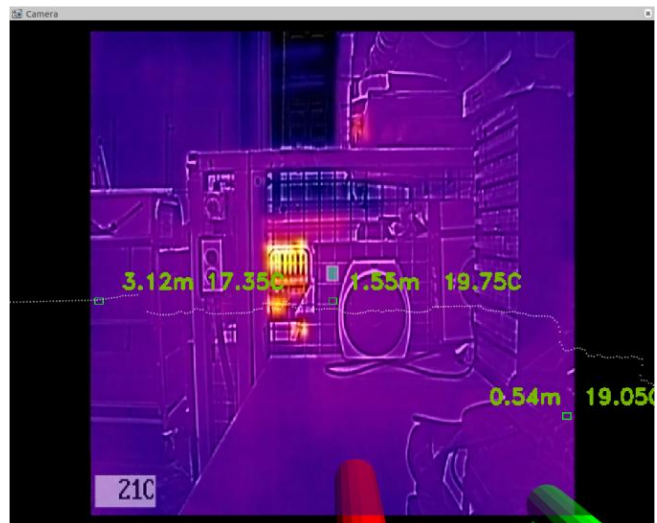

a)

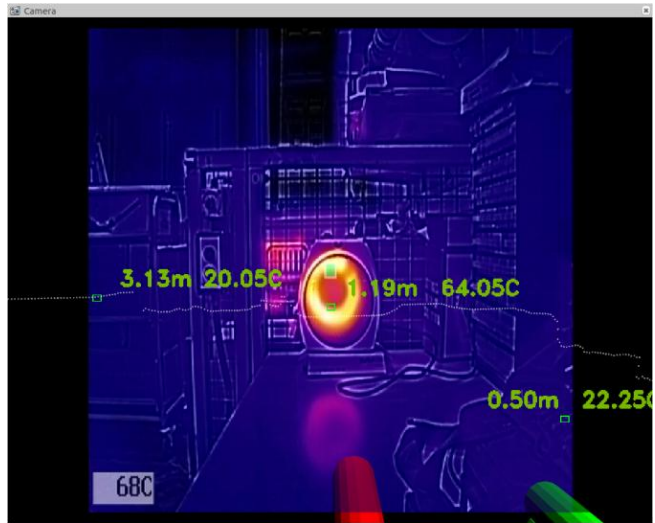
b)

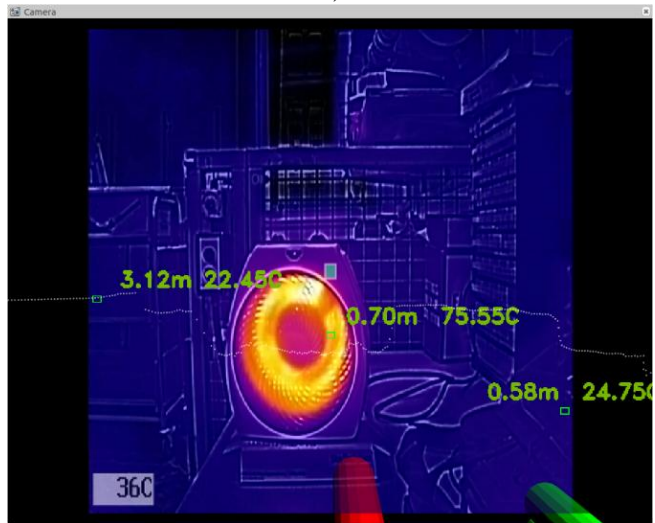
c)

**Fig.12.** Screen captures of thermal images and laser measurements published in RViz and implemented in ROS (Ubuntu): a) combination of three 2D laser readings superimposed on the thermal image; b) combination of three 2D laser readings superimposed on the thermal image when a hot object is located at 1.19m distance (64.05C); c) combination of three 2D laser readings superimposed on the thermal image when a hot object is located at 0.70m distance (75.55C).

In Fig.12 there are shown screen captures of the real-time thermal images and laser measurements published in RViz and implemented in ROS. In Fig.12a is shown a combination of three 2D laser readings superimposed on the thermal image. In Fig.12b is shown a combination of three 2D laser readings superimposed on the thermal image when a hot object is located at 1.19m distance (64.05C). In Fig.12c is shown a combination of three 2D laser readings superimposed on the thermal image when a hot object is located at 0.70m distance (75.55Celsius).

## 6. FUSION OF 3D DEPTH LASER DATA AND THERMAL IMAGE

For measuring the distances to the hotspots it can be used also a 3D MultiSense SLB laser and visualizing the results with the RViz software application which is again a 3D visualization tool for ROS and under Ubuntu/Linux operating system. The $x_h$, $y_h$, $z_h$ positions of a point in space can be obtained by using a 3D MultiSense SLB laser, which is laser, 3D stereo and video. Such a laser can be used in a large number of applications such as automation, autonomous vehicles, 3D mapping, workspace understanding.

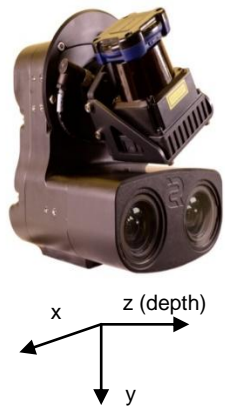

**Fig.13.** MultiSense SLB laser.

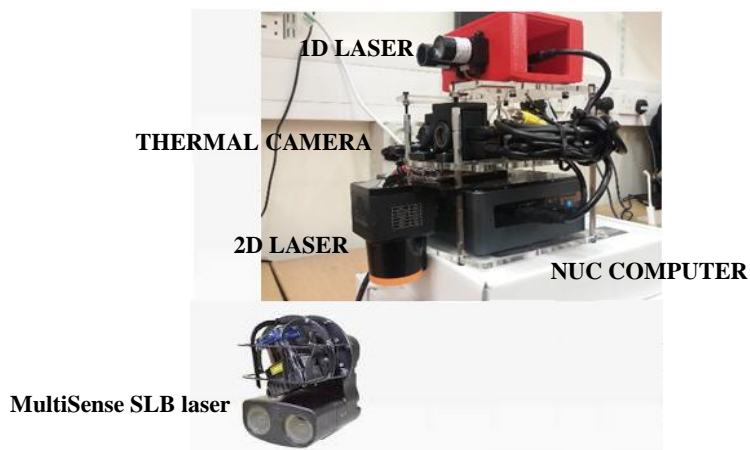

**Fig.14.** Sensor package with 3D MultiSense SLB laser.

It is used the same idea as for the 2D Depth laser and the real-time thermal images/videos and implemented in ROS and visualized in Rviz below.

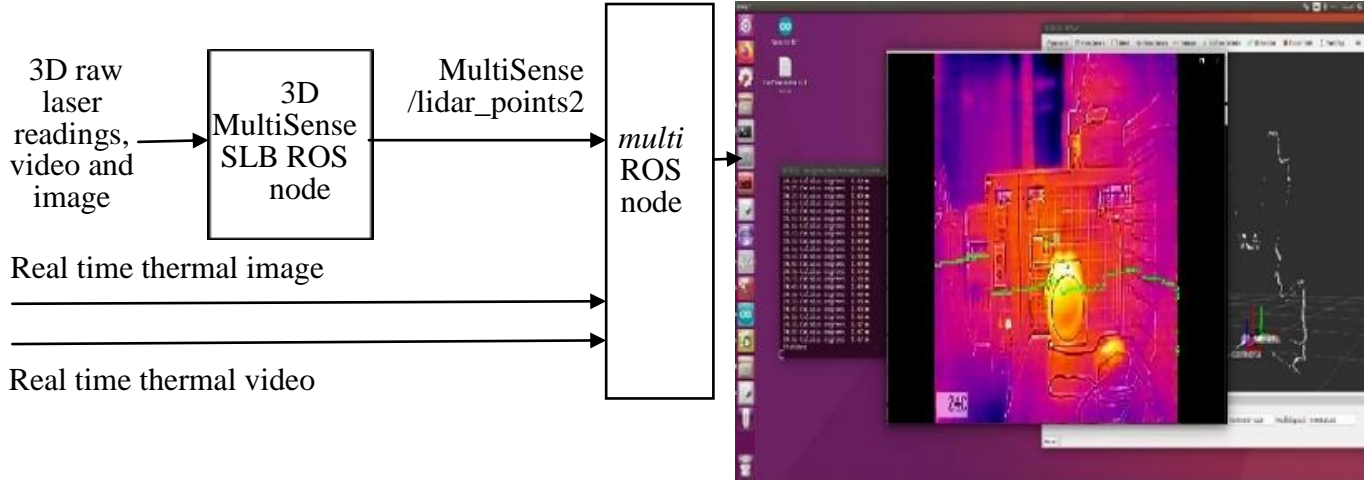

**Fig.15.** ROS nodes for processing of 3D laser readings, real time thermal video and real time thermal image for hotspots identification and localization.

From the 3D MultiSense SLB ROS node there are published a large number of ROS topics (e.g. over 20) and the MultiSense/lidar_points2 topics is used and it contains the $x_h$, $y_h$, $z_h$ positions of a point in space and the laser intensity.

A novel ROS node is written (e.g. called *multi*) which combines the $x_h$, $y_h$, $z_h$ positions of a point in space with the real-time thermal video and thermal images. The same equations as before (i.e. the pinhole camera model) are used here as well:

$$T_{Celsius} = Thermal\_image_{Celsius}\ (x_{pixel}, y_{pixel}) \qquad (7)$$

$$x_{pixel} = c_x + f_x * (x_h / d) \qquad (8)$$

$$y_{pixel} = c_y + f_y * (y_h / d) \qquad (9)$$

Laser points with intensity less than a threshold (e.g. 100) are excluded from recording as well as points with $x_{pixel}$ and $y_{pixel}$ which fall outside of the video/image area (e.g. negative values for $x_{pixel}$ and $y_{pixel}$). The coordinates of a point in space ($x_h$, $y_h$, $z_h$=d) are obtained by using the 3D MultiSense SLB laser device. Distance from the 3D MultiSense SLB laser to the point(s) in space can also be calculated:

$$D = \sqrt{x_h^2 + y_h^2 + z_h^2} \qquad (10)$$

In Fig.16 is shown a comparison between a real-time thermal video obtained in ROS (Fig.16a) and a thermal video with the 3D MultiSense SLB laser data superimposed on the real-time thermal video (Fig.16b-d).

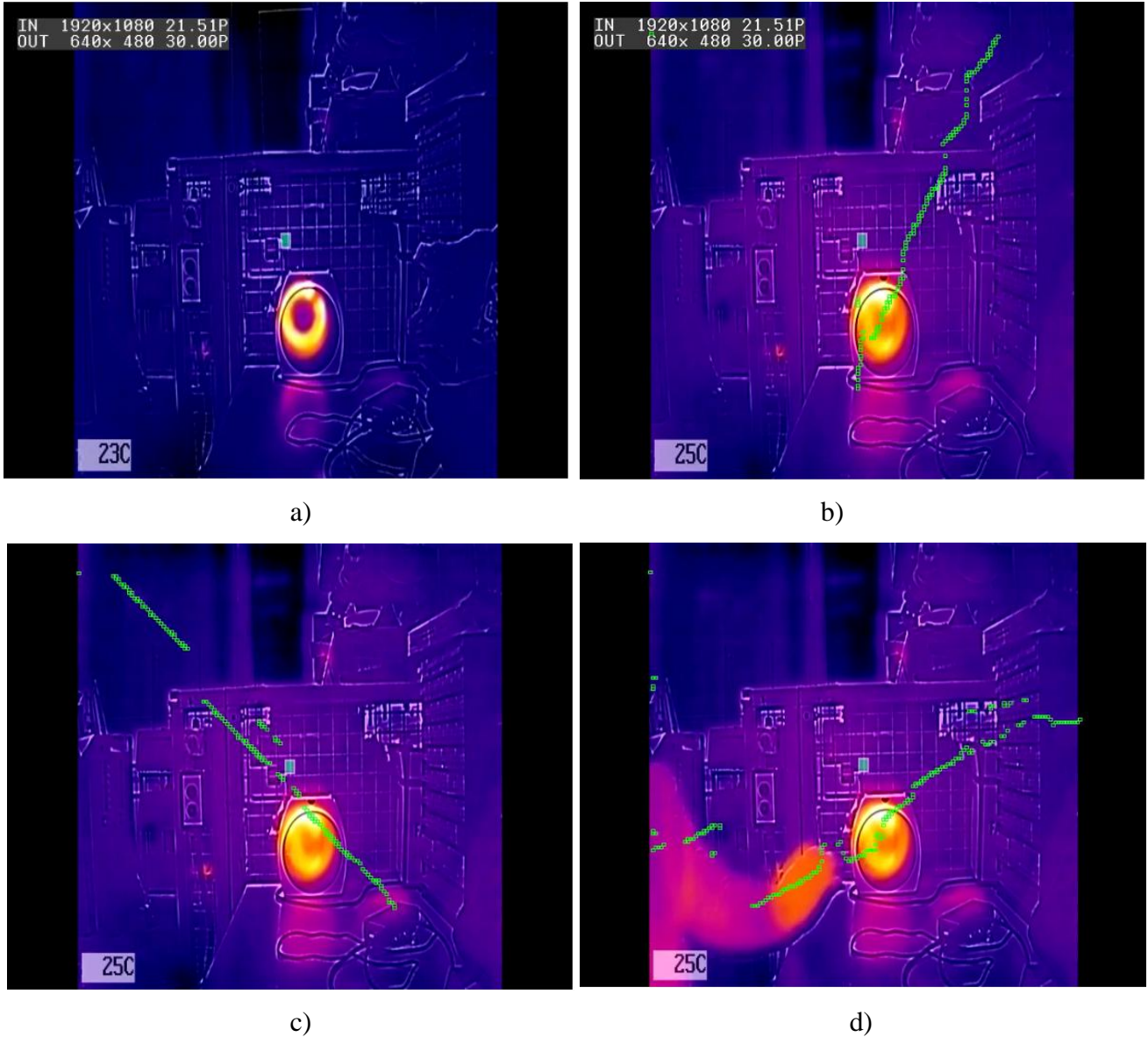

a)  b)  c)  d)

**Fig.16.** Comparison between the thermal real-time video and the thermal real-time video combined with the 3D MultiSense SLB laser data: a) real-time thermal video obtained in ROS; b) combined real-time thermal video with the 3D MultiSense SLB laser data superimposed on it; c) combined real-time thermal video with the 3D MultiSense SLB laser data superimposed on it; d) combined real-time thermal video with the 3D MultiSense SLB laser data superimposed on it.

The temperatures ($T_{Celsius}$ in degrees Celsius), the pixel values ($x_{pixel}$, $y_{pixel}$), the coordinates ($x_h$, $y_h$, $z_h$=d) and the distances (*D*) to all the points in space are also saved in a text file in real-time and can be processed as such by a UAV or robot.

# 7. DISCUSSION AND CONCLUSIONS

This paper develops fast procedures for detection of hotspots in industrial plants such as for example of overheated valves in HDVC stations, and based on operation of UAVs or robots inside the plant/station. For measuring the distances there are used a 1D LIDAR laser, a 2D dimensional Hokuyo laser or a 3D MultiSense SLB laser. A FLIR Duo R dual-sensor thermal camera containing a visible camera and a thermal camera is mounted on a UAV (Figure 2d) and/or a robot for acquisition in real time of thermal videos and thermal images. Previous reported works in the literature took into consideration various other combinations of depth sensing devices and thermal and visible cameras. The results show that the presented methods are robust and the errors produced are small, in the order of millimeters. These methods are suitable for detection and localization of hotspots in real life situations (e.g. industrial plants, HVDC electrical substations, environment [37-40] and agriculture [41-42]) where the use of UAVs or robots [43-46] is needed.

Future work will involve more lab testing and field work as well as testing in more real life situations. Other work will involve also taken into consideration the distortion of the lenses, which was addressed already in the literature below, and the angles of rotation of UAVs. Furthermore, the work will address even more simple technical solutions which will be able to provide similar support as above: for example with the scope of decreasing the additional weight of the computing platform (i.e. NUC computer) and increasing the flight autonomy of the UAVs or the life autonomy of robot (e.g. use of Raspberry Pi boards).

# 8. ACKNOWLEDGMENTS

This work was supported by the HOME Offshore project (EPSRC UK No. EP/P009743/1). The authors would like to thank to Dr. Simon Watson, Dr. Liam Brown, Dr. Andrew West, Prof Mike Barnes, Mr. Kristopher Kabbabe and Prof. William Crowther for supporting this work.